\documentclass[sigconf]{acmart}

\setcopyright{none}
\renewcommand\footnotetextcopyrightpermission[1]{}
\usepackage{graphicx}
\usepackage{microtype}
\acmConference[XAI for U '26]{3rd International Workshop on Explainable AI for Ubiquitous, Pervasive and Wearable Computing}{October 11--12, 2026}{Shanghai, China}
\acmYear{2026}

\title{TrustmeWatcher: An Application for Workplace Micro-Sensing and Explainable Well-Being Feedback}

\author{Chengyu Yu}
\affiliation{%
  \institution{Università della Svizzera italiana (USI)}
  \city{Lugano}
  \country{Switzerland}
}

\author{Leon Jacopo Costa}
\affiliation{%
  \institution{Università della Svizzera italiana (USI)}
  \city{Lugano}
  \country{Switzerland}
}

\author{Zoja Anžur}
\affiliation{%
  \institution{Jožef Stefan Institute (JSI)}
  \city{Ljubljana}
  \country{Slovenia}
}
\additionalaffiliation{%
  \institution{Jožef Stefan International Postgraduate School}
  \city{Ljubljana}
  \country{Slovenia}
}

\author{Mohan Li}
\affiliation{%
  \institution{Università della Svizzera italiana (USI)}
  \city{Lugano}
  \country{Switzerland}
}

\author{Gašper Slapničar}
\authornotemark[1]
\affiliation{%
  \institution{Jožef Stefan Institute (JSI)}
  \city{Ljubljana}
  \country{Slovenia}
}

\author{Daniil Kirilenko}
\affiliation{%
  \institution{Università della Svizzera italiana (USI)}
  \city{Lugano}
  \country{Switzerland}
}

\author{Martin Gjoreski}
\affiliation{%
  \institution{Università della Svizzera italiana (USI)}
  \city{Lugano}
  \country{Switzerland}
}

\author{Mitja Luštrek}
\authornotemark[1]
\affiliation{%
  \institution{Jožef Stefan Institute (JSI)}
  \city{Ljubljana}
  \country{Slovenia}
}

\author{Marc Langheinrich}
\affiliation{%
  \institution{Università della Svizzera italiana (USI)}
  \city{Lugano}
  \country{Switzerland}
}

\begin{document}

\begin{abstract}
Workplace sensing studies combine long-running behaviour traces with self-reports, yet the tools that collect those data often sit apart from the interface that returns results. We present TrustmeWatcher, the application built for the TRUST-ME project to connect this work. TrustmeWatcher reuses ActivityWatch's OS-level watchers for computer-activity collection and adds its own application layer. It turns the collected traces into an interactive screen-time dashboard, synchronizes responses from short questionnaires completed on the StreamDeck, and presents questionnaires alongside video highlights. Activity records and self-reports are aligned into labelled records for model development. The scope of this paper is limited to ActivityWatch data as model input. Artificial intelligence (AI) uses these activity records to predict six normalized state scores and an overall well-being score. The trained model runs locally, and the dashboard presents its predictions in semantic bands. Explainable artificial intelligence (XAI) helps users understand how recorded activity contributed to a prediction. Privacy Control lets users pause or resume the camera and eye tracker used by the study. We describe the workflow, its user-device and sensing-setup boundaries, and its use with records from 17 participants. The result is a deployed application and study workflow that integrates activity review, study data collection, privacy control, local prediction, and a participant-facing interface for XAI evaluation.
\end{abstract}

\ccsdesc[500]{Human-centered computing~Ubiquitous and mobile computing}
\ccsdesc[300]{Computing methodologies~Machine learning}

\keywords{explainable AI, workplace well-being, micro-sensing, self-report, privacy control, ubiquitous computing}

\maketitle

\section{Introduction}

Digital traces can reveal work patterns that repeated surveys alone miss, including changes in engagement, fatigue, stress, physical comfort, and social interaction \cite{anzur2025wellbeing}. The same traces can become instruments of workplace surveillance. TRUST-ME (TRUstworthy enhancement of job SaTisfaction and productivity using Micro-sensing in work Environments) brings together Università della Svizzera italiana and the Jo\v{z}ef Stefan Institute to study this conflict \cite{trustme2023project}. Its aim is to examine how sensing and AI can support knowledge workers while preserving privacy, autonomy and providing understandable explanations for its predictions.

\begin{figure}[t]
  \centering
  \includegraphics[width=0.90\columnwidth]{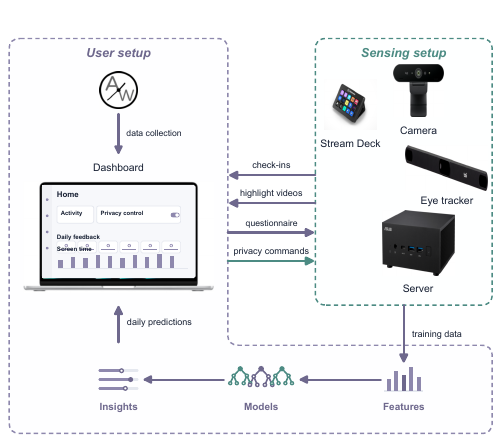}
  \Description{The diagram separates the participant's computer, the sensing setup, and an external model development pipeline. The sensing setup sends check-ins and video highlights to the participant's computer, while questionnaire responses are stored on the study server. Privacy commands from the participant control the camera and eye tracker. Study data are used to train the model, which is then deployed to the participant's computer for local prediction.}
  \caption{TrustmeWatcher system overview and data flow.}
  \label{fig:architecture}
\end{figure}

Computer activity and self-reports describe different perspectives of a workday. Activity logs record how the computer was used, but they do not say whether a person felt focused, tired, stressed, or productive. Self-reports capture that experience, although they are subjective and burdensome if requested too often. Automatically measured and self-reported productivity can diverge, so neither should stand in for the other \cite{beller2021productivity}. The study collects two forms of self-report: periodic nine-question check-ins using the Stream Deck device seen in Figure~\ref{fig:architecture} during work, and video questionnaires completed after reviewing recorded highlights. TRUST-ME also uses data collected through ActivityWatch namely application usage, mouse and keyboard input. A camera, an eye tracker, audio, and recorded highlights are also used to provide study context. Because the camera and eye tracker capture sensitive contextual data, participants need a visible and reversible way to pause or resume sensing. Such control is particularly important in workplace settings, where power asymmetries can make consent less freely given \cite{chowdhary2023consent}. It may also help reduce concerns about surveillance, which is associated with lower autonomy and poorer worker well-being \cite{glavin2024surveillance}.

The TRUST-ME study has two phases. The first collected computer activity and repeated self-reports from 17 participants for model development and evaluation. The second, currently being prepared, will use TrustmeWatcher as its participant-facing application. For computer-activity collection, TrustmeWatcher reuses ActivityWatch’s open-source watchers. These background processes locally record application usage and keyboard and mouse activity \cite{activitywatch2024}.
TrustmeWatcher's own service and interface turn the resulting traces into an interactive screen-time dashboard. They also synchronize responses from both questionnaires, prepare labeled records for model development, and run the trained model locally. Daily feedback and ambient-sensor controls use the same interface. Figure~\ref{fig:architecture} shows the boundaries between the user device, the sensing setup, and the modeling pipeline.

This paper contributes:
\begin{itemize}
  \item an integrated application with an interactive screen-time dashboard that lets participants explore computer activity from hours to years, alongside self-report collection, privacy controls, and daily well-being and productivity predictions;
  \item a longitudinal field-study record that connects repeated self-reports from 17 participants with the computer activity surrounding them; and
  \item an implemented participant-facing counterfactual interface for examining explanation use across repeated encounters.
\end{itemize}

\section{Related Work}

Ubiquitous-computing platforms often combine passive sensing with active reports. StudentLife \cite{wang2014studentlife} paired continuous phone sensing with ecological momentary assessment over a university term, while deliberately withholding feedback to avoid changing participant behaviour. RADAR-base supports passive phone and wearable sensing together with active questionnaires, study management, storage, and visualization \cite{ranjan2018radar}.
In the second phase, TrustmeWatcher gives the participant device a different role. The person who produces the data can review the recorded activity, control ambient sensing, and receive model estimates and explanations in the same application. This makes those interactions observable parts of the participant's study experience.

Device interaction, physiological sensing, environmental data, and self-reports have different validity and privacy costs \cite{anzur2025wellbeing}. Beller et al. paired desktop telemetry with daily productivity ratings and found that activity measures cannot replace a person's account of the day \cite{beller2021productivity}.
TrustmeWatcher aligns both forms of self-report with the surrounding computer activity.

The model layer generates counterfactual results, while TrustmeWatcher presents them through its participant-facing interface. BayCon produces model-agnostic counterfactual instances over mixed features \cite{romashov2022baycon}. Cheruvalath et al. combine attribution, counterfactuals, and accessible text for non-expert users \cite{cheruvalath2025explanations}. The generated explanation still has to fit the context and questions that matter to its user \cite{miller2019explanation}. Repeated use adds questions about comprehension, trust calibration,
and changing mental models that a single encounter cannot answer.

\section{TrustmeWatcher Application}

\subsection{System Roles and Data Flow}

\begin{figure}[b]
  \centering
  \includegraphics[width=0.82\columnwidth]{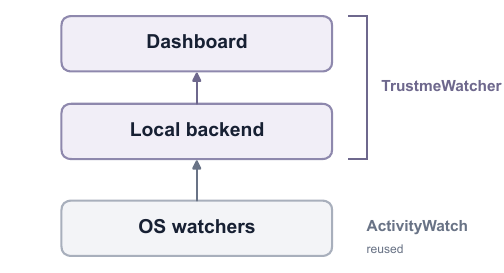}
  \Description{Three local software layers are stacked vertically. Reused ActivityWatch OS watchers form the bottom layer. The TrustmeWatcher local backend and dashboard form the two layers above it.}
  \caption{Architecture view.}
  \label{fig:software-layers}
\end{figure}

TrustmeWatcher separates reused OS-level ActivityWatch watchers from its own local backend and dashboard (Figure~\ref{fig:software-layers}). ActivityWatch watchers record computer activity on the user device. The local backend turns those events into a unified activity history and supports the study-data synchronization, Privacy Control, and local inference paths described below.

The sensing setup provides check-ins, video highlights, and ambient sensing. TrustmeWatcher brings these inputs into the same user workflow, returns video questionnaire responses to the study server, and carries Privacy Control commands back to the camera and eye tracker.

Model development sits outside both runtime boundaries. The study records combine activity context with both forms of self-report for training and evaluation. The trained model is then deployed to the user device. TrustmeWatcher prepares features from the local activity history, runs the model locally, and stores one daily report for the interface. Training data and model predictions follow different routes, as shown in Figure~\ref{fig:architecture}.

\subsection{A Unified Workplace Study Workflow}

\begin{figure}[t]
  \centering
  \includegraphics[width=\columnwidth,pagebox=cropbox]{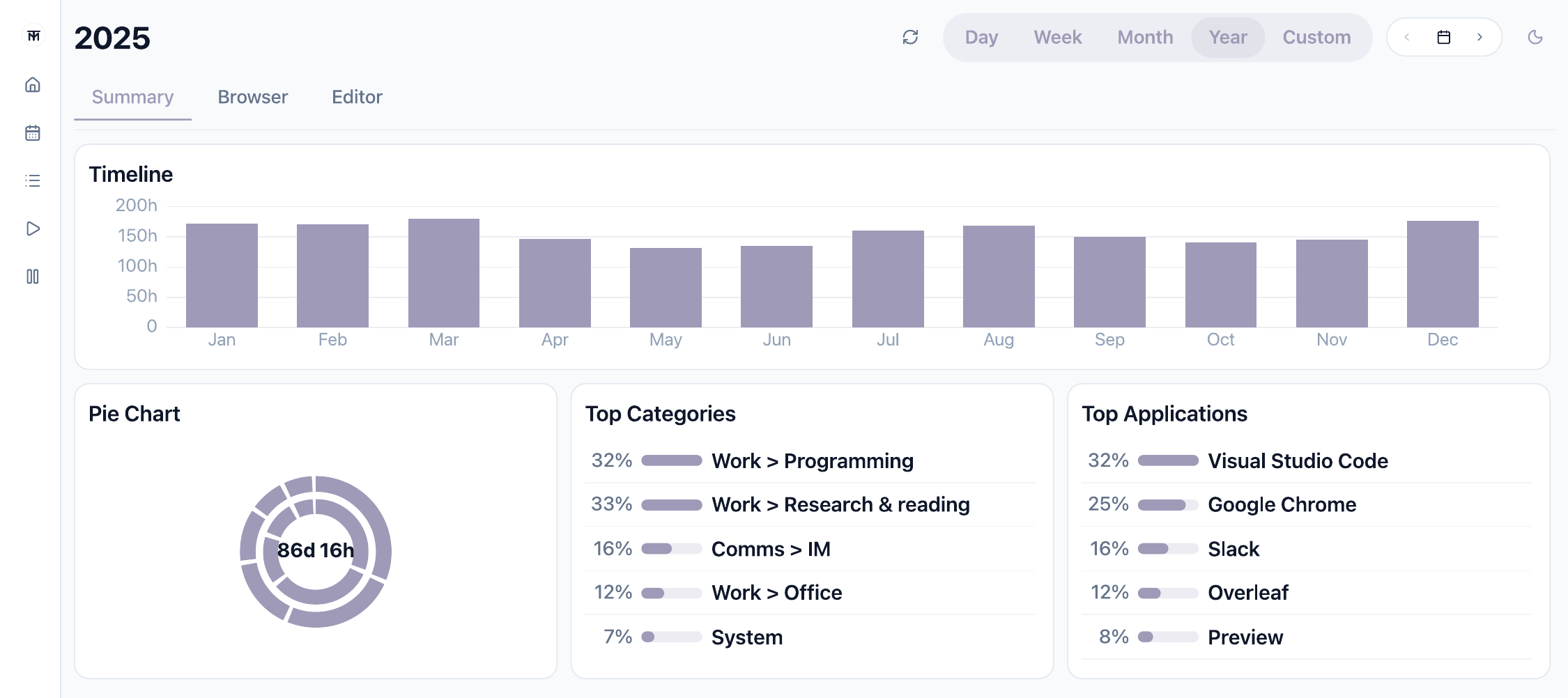}
  \Description{The complete screen-time summary showing a yearly timeline, category distribution, top categories, and top applications.}
  \caption{Screen-time summary using synthetic data.}
  \label{fig:interfaces}
\end{figure}

TrustmeWatcher presents computer activity through a screen-time interface for immediate review and longer-range exploration (Figure~\ref{fig:interfaces}). Users can move from hourly activity to weekly, monthly, yearly, and custom periods without changing how applications and categories are defined. The summary combines a timeline, an application distribution, and ranked category and application lists. This gives users a readable account of the same activity history that supports the study.

The application also presents check-ins and video highlights. Participants complete check-ins during work. The application groups the available highlights for a day, allowing the user to review several relevant moments. The two channels provide different temporal views of the same work experience.

\begin{figure}[t]
  \centering
  \includegraphics[width=\columnwidth,pagebox=cropbox]{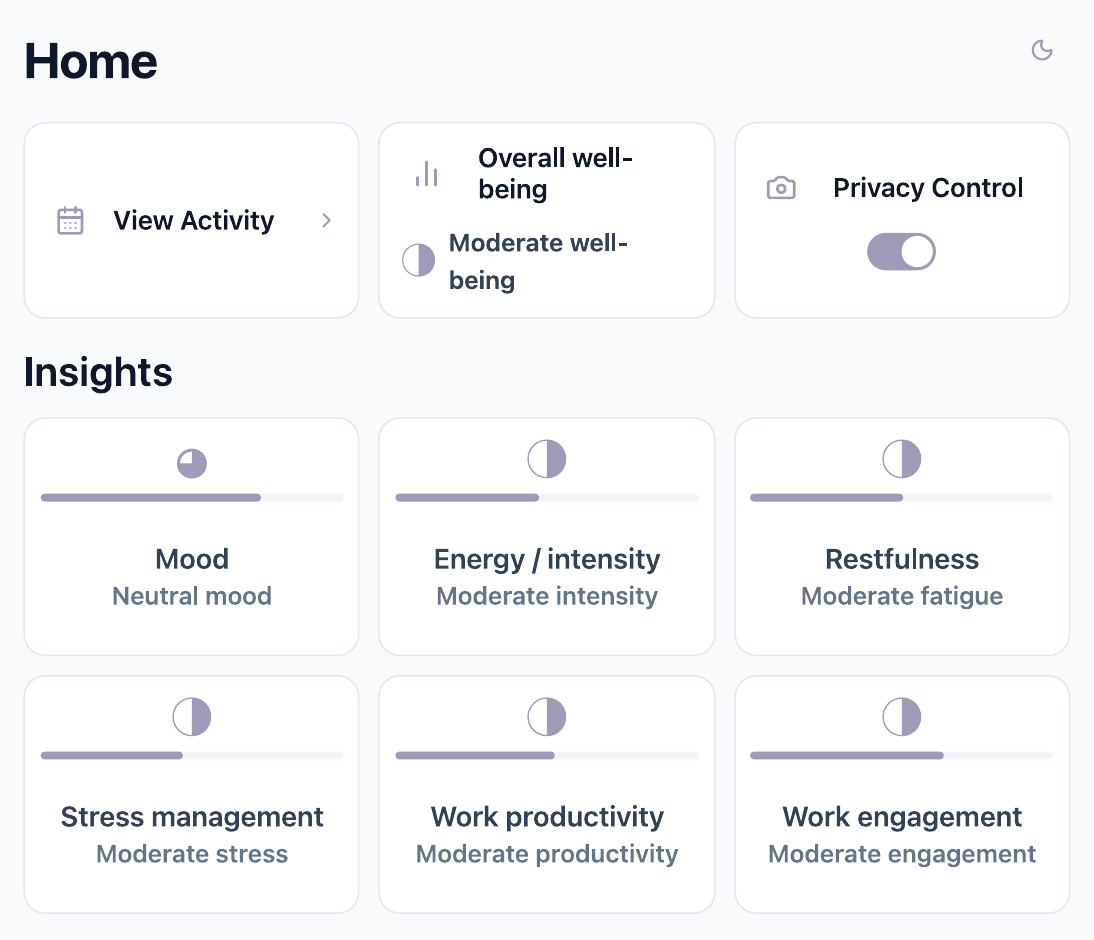}
  \Description{The user dashboard Home view showing daily semantic feedback cards beside direct control of the study sensors.}
  \caption{Daily feedback and sensing control using synthetic data.}
  \label{fig:home-interface}
\end{figure}

TrustmeWatcher integrates Privacy Control into the application (Figure~\ref{fig:home-interface}). Users can see whether camera and eye-tracker sensing is active and pause or resume it. Pausing Privacy Control turns off the camera and eye tracker. Placing this control beside returned feedback makes consent an ongoing choice instead of an installation-time setting \cite{chowdhary2023consent}.

\subsection{From Study Signals to Model Estimates}

The study collects nine responses on a Likert scale: mood (\(q_1\)), 
\([-3,+3]\); emotion intensity (\(q_2\)), \([0,6]\); tiredness 
(\(q_3\)), \([0,6]\); work enthusiasm (\(q_4\)), \([0,6]\); 
work immersion (\(q_5\)), \([0,6]\); physical comfort (\(q_6\)), 
\([-3,+3]\); social experience (\(q_7\)), \([-3,+3]\); work stress 
(\(q_8\)), \([0,6]\); and productivity (\(q_9\)), \([0,6]\).
In the first phase, participants answered the same questions in check-ins during work and in video questionnaires after reviewing recorded highlights. Responses from both sources were combined for model training.

The prediction pipeline uses seven regressors to estimate six state scores 
and overall well-being. The training targets are calculated from the 
questionnaire responses as mood valence \(V=q_1+3\), arousal \(A=q_2\), 
restfulness \(R=6-q_3\), stress management \(S=6-q_8\), productivity 
\(P=q_9\), engagement \(E=(q_4+q_5)/2\), and overall well-being 
\(W=(V+E+S)/3\). Tiredness and stress are reversed so higher scores always 
indicate a better state. Physical comfort (\(q_6\)) and social experience 
(\(q_7\)) are collected but are not prediction targets. 

The target construction follows the questionnaire design: distinct states are modeled separately, while complementary items are averaged into composite scores. Cronbach’s $\alpha$ was used to assess which state scores were consistent enough to combine for overall well-being. Based on this, valence, engagement, and stress management were included, while fatigue and arousal were excluded.
For each target multiple regressors use the same 25-feature vector. This vector summarizes the preceding 60 minutes together with 24-hour and seven-day context. The features describe learned activity-state shares, application-category ratios, focus fragmentation, keyboard and pointer activity, and changes from recent history. Features are standardized separately for each participant using statistics fitted on training data only. A participant not present during training uses global training-set statistics. For each target, the regressor with the lowest validation MSE is selected. 

During inference, the same feature pipeline constructs the input vector locally from ActivityWatch data. The application presents the seven estimates on the common 0--6 scale.

\subsection{From Model Estimates to Explanations}

\begin{figure}[!htbp]
  \centering
  \includegraphics[width=0.70\columnwidth,pagebox=cropbox]{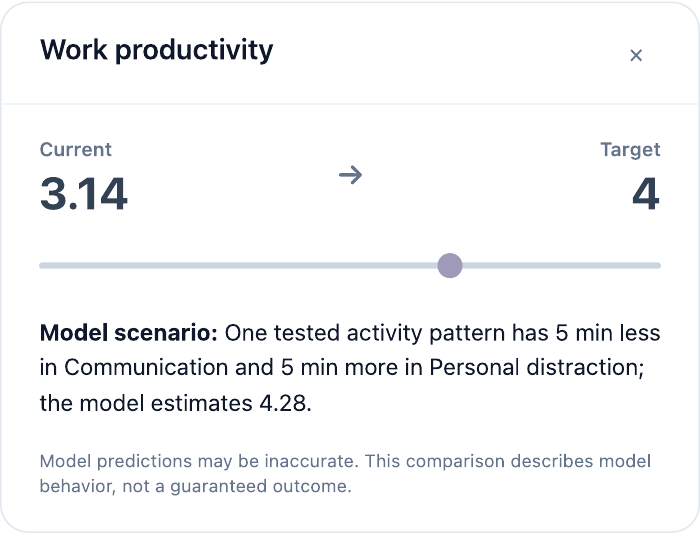}
  \Description{A counterfactual dialog for work productivity. It shows a current score of 3.14, a selected target of 4, a slider, and one model scenario that reallocates five minutes from communication to personal distraction, producing a model estimate of 4.28. A short disclaimer states that model predictions may be inaccurate and do not guarantee an outcome.}
  \caption{Counterfactual example.}
  \label{fig:counterfactual}
\end{figure}

TrustmeWatcher presents the six state scores as semantic cards in Figure~\ref{fig:home-interface}. Scores below 2 map to the low band, scores from 2 to below 5 to moderate, and scores from 5 to 6 to high. Labels follow each construct: a high stress-management score appears as low stress, and a high restfulness score as low fatigue. Selecting a card opens the counterfactual view in Figure~\ref{fig:counterfactual}.

The user chooses a target score on the shared 0--6 scale. Starting from the current 60-minute computer-activity allocation, the search swaps time between personal distraction, media, communication, development, writing and research in five-minute units. A swap removes five minutes from one category and adds those five minutes to another category. The system evaluates every valid allocation and returns the one whose model estimate is closest to the selected target. This zero-sum search adopts the framing of counterfactual explanation used in prior work \cite{romashov2022baycon,cheruvalath2025explanations}.

\section{Field Study and Model Evaluation}

\subsection{Data-Collection Study}

\paragraph{Participants and setting.}
The TRUST-ME data-collection study followed knowledge workers in their everyday work setting rather than in a single laboratory session. The current archive contains pseudonymized records from 17 participants and spans July 2025 to March 2026. The amount of recorded self-report data varies with each participant's work schedule: individual records cover between 10 and 28 distinct workdays, with a median of 18.

\paragraph{Procedure.}
Computer activity was recorded in the background while participants worked. At intervals during the day, participants completed check-ins on the StreamDeck to describe their current state. The sensing setup also produced video highlights for retrospective reflection. For each available highlight, participants rated the same dimensions in the video questionnaire while viewing the recorded moment. This produced two kinds of self-report: an immediate response close to the experience and a later response grounded in a concrete episode from the day.

\paragraph{Study record.}
All 17 participants contributed computer-activity context and check-ins; nine also contributed video questionnaire responses. TrustmeWatcher preserves the response source and timestamp when it aligns each self-report with the preceding activity context. This allows the two channels to be analysed together without treating them as interchangeable labels. The aligned records feed the external model-development pipeline, and the deployed model later uses recent and longer-range activity features to produce daily estimates on the user's device.

\subsection{Prediction Performance}

Questionnaire days were ordered chronologically within each participant and divided approximately 80/10/10 into training, validation, and test sets. One complete day at each boundary was excluded to reduce temporal overlap. Instances were not shuffled to avoid data-leakage. Models were selected using validation MSE and evaluated once on the held-out test set.

We compared the model with random, global-mean, and personal-mean baselines. The random baseline draws scores uniformly from the 0--6 response scale. The global mean is calculated separately for each target using the training and validation responses. The personal mean is calculated from each participant's own training and validation responses.

\begin{table}[!htbp]
  \caption{Within-user prediction performance}
  \label{tab:prediction-performance}
  \centering
  \scriptsize
  \setlength{\tabcolsep}{5pt}
  \begin{tabular}{@{}lrr@{}}
    \toprule
    Model & Validation MSE $\downarrow$ & Test MSE $\downarrow$ \\
    \midrule
    Random baseline & $5.9 $ & $5.8$ \\
    Global-mean baseline & 1.4 & 1.3 \\
    Personal-mean baseline & 1.0 & 0.85 \\
    Our model & \textbf{0.85} & \textbf{0.78} \\
    \bottomrule
  \end{tabular}
\end{table}

The global mean tests improvement over a population-level constant. The personal mean provides a stricter comparison because it tests whether the activity features add information beyond each participant's typical response.

\section{Discussion}

The reported evaluation covers prediction performance and data-collection coverage; explanation comprehension, trust calibration, and interpretation of model scenarios require analysis of repeated participant interactions \cite{gjoreski2024longitudinal}. Computer activity omits meetings and off-screen work, and application categories cannot establish intent, so estimates reflect recorded context rather than complete accounts of work or well-being.

\section{Conclusion}

TrustmeWatcher integrates a model developed from records collected from 17 participants with computer-activity review, self-report, privacy control, local prediction, and counterfactual explanations for the second study phase. It provides the infrastructure for longitudinal evaluation of explanation use, trust, and privacy choices.

\begin{acks}
This work was funded by SNSF through the projects TRUST-ME (GA no. 214991) and XAI-PAC (GA no. 216405).
\end{acks}

\enlargethispage{\baselineskip}\bibliographystyle{ACM-Reference-Format}
\bibliography{references}

\end{document}